\documentclass[runningheads,a4paper]{llncs}

\usepackage{amssymb}
\usepackage{graphicx}

\usepackage{bm}
\usepackage{amsmath}
\usepackage{commath}
\usepackage{float}
\usepackage{blindtext}
\usepackage{scrextend}

\usepackage{url}
\urldef{\mailsa}\path|jluo5@bwh.harvard.edu|    
\newcommand{\keywords}[1]{\par\addvspace\baselineskip
\noindent\keywordname\enspace\ignorespaces#1}

\begin{document}

\mainmatter  

\title{A Feature-Driven Active Framework for Ultrasound-Based Brain Shift Compensation}
\vspace{-6mm}
\titlerunning{\scriptsize{ A Feature-Driven Active Framework for US-based Brain Shift Compensation}}

%

\author{Jie Luo$^{1,2}$,  Matt Toews$^{3}$,  Ines Machado$^{1}$, Sarah Frisken$^{1}$, Miaomiao Zhang$^{4}$, Frank Preiswerk$^{1}$, Alireza Sedghi$^{1}$, Hongyi Ding$^5$, Steve Pieper$^{1}$, Polina Golland$^{6}$, Alexandra Golby$^{1}$, Masashi Sugiyama$^{7,2}$ and William M. Wells III$^{1,6}$ }
\authorrunning{J.Luo et al.}

	
\institute{ $^1$Brigham and Women's Hospital, Harvard Medical School, USA\\ $^2$Graduate School of Frontier Sciences, The University of Tokyo, Japan\\ $^3$Ecole de Technologie Superieure, Canada\\ $^4$Computer Science and Engineering Department, Lehigh University, USA\\$^5$Department of Computer Science, The University of Tokyo, Japan \\$^6$Computer Science and Artificial Intelligence Laboratory, MIT, USA \\$^7$Center for Advanced Intelligence Project, RIKEN, Japan \\
	\mailsa 
}

%
%

\toctitle{Lecture Notes in Computer Science}
\tocauthor{Authors' Instructions}
\maketitle

\vspace{-6mm}
\begin{abstract}
\emph{A reliable Ultrasound (US)-to-US registration method to compensate for brain shift would substantially improve Image-Guided Neurological Surgery.  Developing such a registration method is very challenging, due to factors such as missing correspondence in images, the complexity of brain pathology and the demand for fast computation. We propose a novel feature-driven active framework. Here, landmarks and their displacement are first estimated from a pair of US images using corresponding local image features.  Subsequently, a Gaussian Process (GP) model is used to interpolate a dense deformation field from the sparse landmarks. Kernels of the GP are estimated by using variograms and a discrete grid search method. If necessary, the user can actively add new landmarks based on the image context and visualization of the uncertainty measure provided by the GP to further improve the result. We retrospectively demonstrate our registration framework as a robust and accurate brain shift compensation solution on clinical data acquired during neurosurgery. }

\end{abstract}

\vspace{-9mm}
\section{Introduction}

Surgical resection is the initial treatment for nearly all brain tumors. The achieved extent-of-resection is strongly correlated with prognosis and is the single greatest modifiable determinant of survival. Brain tumors are intimately involved in surrounding functioning brain tissue, aggressive resection must be balanced against the risk of causing new neurological deficits. 

During neurosurgery, Image-Guided Neurosurgical Systems (IGNSs) provide a patient-to-image mapping that relates the preoperative image data to an intraoperative patient coordinate system, allowing surgeons to infer the locations of their surgical instruments relative to preoperative image data and helping them to optimize the extent of resection while avoiding damage to critical structures.

Commercial IGNSs assume a rigid registration between preoperative imaging and patient coordinates. However, intraoperative deformation of the brain, which is also known as brain shift, invalidates this assumption. Since brain shift progresses during surgery, the rigid patient-to-image mapping of IGNS becomes less and less accurate. Consequently, most surgeons only use IGNS to make a surgical plan but justifiably do not trust it throughout the course of an operation \cite{Gerard,Bayer}. 
 
 \subsubsection{Related Work}
 As one of the most important error sources in IGNS, intraoperative brain shift must be compensated in order to increase the accuracy of neurosurgery. Registration between the Intraoperative MRI (iMRI) image, which provides clinicians with an updated view of anatomy during surgery, and preoperative MRI (preMRI) image (preop-to-intraop registration) has been a successful strategy for brain shift compensation \cite{Hata,Soza,Clatz,Vigneron,Drakopoulos}. However, iMRI acquisition is disruptive, expensive and time consuming, making this technology unavailable for most clinical centers worldwide. More recently, 3D intraoperative Ultrasound (iUS) appears to be a promising replacement for iMRI. Although some progress has been made by previous work on preMRI-to-iUS registration \cite{Gobbi,Arbel,Pennec,Lette,Reinerstsen,Fuerst,Rivaz}, yet there are still no clinically accepted solutions and no commercial neuro-navigation systems that provide brain shift compensation. This is because of three reasons: 1) Most non-rigid registration methods can not handle artifacts and missing structures in iUS; 2) The multi-modality of preMRI-to-iUS registration makes the already difficult problem even more challenging; 3) A few methods \cite{Ou} can achieve a reasonable alignment, yet they take around 50 minutes for an US pair and are too slow to be clinically applicable.
 
 Another shortcoming of existing brain shift compensation approaches is the lack of an uncertainty measure. Brain shift is a complex spatiotemporal phenomenon, and given the state of registration technology, and the importance of the result, it seems reasonable to expect, e.g., error bars that indicate the confidence level in the estimated deformation. In fact, registration uncertainty can actually helps surgeons make more informed decisions. If a surgeon must decide whether to continue resection near a critical structure, it is vital that they know how far the instrument is predicted to be from the structure and how likely the prediction is to be accurate. Moreover, if a large registration error at location A and small error at location B are observed in the vicinity of surgical field, without knowledge of registration uncertainty, the surgeon would probably assume a large error everywhere and thus ignore the registration altogether. If only s/he knows that A lies in an area of high uncertainty while B lies in an area of low uncertainty, s/he would have greater confidence in the registration at B and other locations of low uncertainty.

 In this paper, we propose a novel feature-driven active framework for brain shift compensation. Here, landmarks and their displacement are first estimated from a pair of US images using corresponding local image features.  Subsequently, a Gaussian Process (GP) model \cite{GPBook} is used to interpolate a dense deformation field from the sparse landmarks. Kernels of the GP are estimated by using variograms and a discrete grid search method. If necessary, for areas that are difficult to align, the user can actively add new landmarks based on the image context and visualization of the uncertainty measure provided by the GP to further improve the registration accuracy.
 
  Contributions and novelties of our work can be summarized as follows:
   \vspace{-2mm}
  \begin{enumerate}
  	\item The proposed feature-based registration is robust for aligning iUS image pairs with missing correspondence and is fast.
    \item We explore applying a GP model and variograms for image registration.
 	\item Registration uncertainty in transformation parameters can be naturally obtained from the GP model.
 	\item To the best of our knowledge, the proposed active registration strategy is the first method to actively combine user expertise in brain shift compensation.
 	\item We retrospectively demonstrate the efficacy of our method on clinical data acquired during neurosurgery.
 \end{enumerate}

\vspace{-6mm}
\section{Method}
\vspace{-2mm}

\subsection{The role of US-to-US registration}
In order to alleviate the difficulty of preop-to-intraop registration, instead of directly aligning iMRI and iUS images, we choose an iterative compensation approach which is similar to the work in \cite{Riva}.

As shown in Fig.1. acquisition processes for pre-duraUS (preUS) and post-resectionUS (postUS) take place before opening the dura and after tumor resection, respectively. Since most brain-shift occurs after taking the preUS, a rigid multi-modal registration may be suffice to achieve a good alignment $T^{\mathrm{rigid}}$ between preMRI and preUS \cite{Fuerst}. Next, we register the preUS to postUS using the proposed feature-driven active framework to acquire a deformable mapping $T_{\mathrm{deform}}$. After propagating $T_{\mathrm{rigid}}$ and $T_{\mathrm{deform}}$ to the preMRI, surgeons may use it as an updated view of anatomy to compensate for brain shift during the surgery.

\vspace{-6mm}
\begin{figure}[H]
	\centering
	\includegraphics[height=2.9cm]{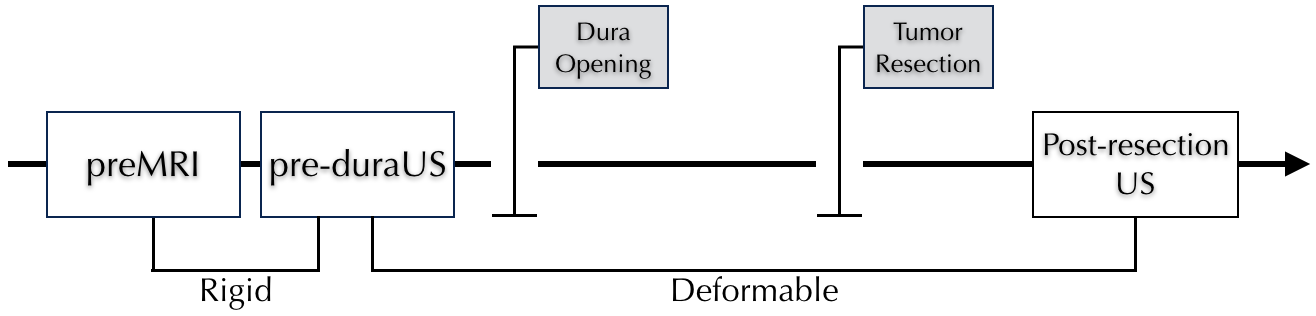}
	\vspace{-2mm}
	\caption{Pipeline of the US-based brain shift compensation.  }
	\label{fig:correct1}
	\vspace{-6mm}
\end{figure}

\subsection{Feature-based registration strategy}

Because of tumor resection, compensating for brain shift requires non-rigid registration algorithms capable of aligning structures in one image that have no correspondences in the other image. In this situation, many image registration methods that take into account the intensity pattern of the entire image will become trapped in incorrect local minima.

\begin{figure}[t]
	\centering
	\includegraphics[height=5.6cm]{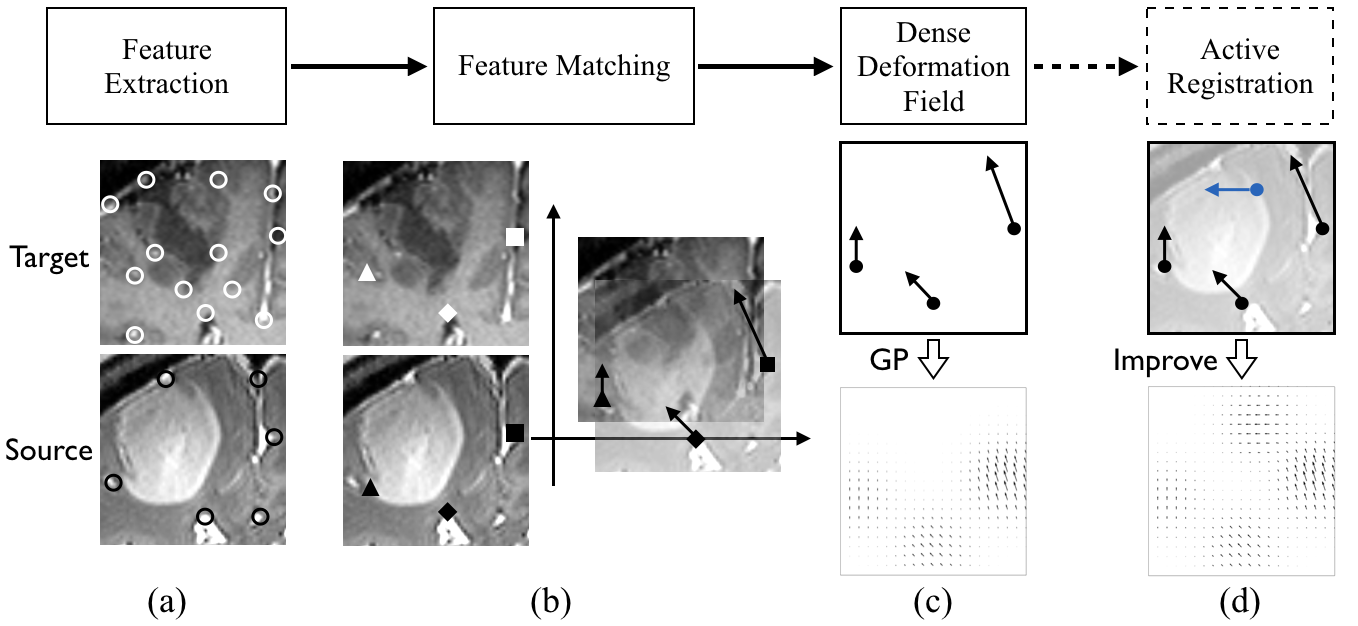}
	\vspace{-7mm}
	\caption{Pipeline of the feature-based active preduraUS-to-postUS registration.}
	\label{fig:correct1}
		\vspace{-3mm}
\end{figure}

We therefore pursue a Feature-Based Registration (FBR) strategy due to its robustness in registering images with missing correspondence \cite{Matt}. FBR mainly consists of 3 steps: feature-extraction, feature-matching and dense deformation field estimation. An optional ``active registration'' step can be added depends on the quality of FBR.

\vspace{-4mm}

\subsubsection{Feature extraction and matching} As illustrated in Fig.2(a)(b), distinctive local image features are automatically extracted and identified as key-points on preUS and postUS images. A matcher searches for a corresponding postUS key-point for each key-point on the preUS image \cite{Matt}. 

From a matched key-point pair, let $\mathbf{x}_i$ be the coordinates of the predUS key-point and $\mathbf{x}^{\mathrm{post}}_i$ be the coordinate of its postUS counterpart. Here, we first use all matched PreUS key-points as landmarks, and perform a land-mark based preUS-to-postUS affine registration to obtain a rough alignment. $\mathbf{x}^{\mathrm{post}}_i$ becomes $\mathbf{x}^{\mathrm{affine}_i}$ after the affine registration. The displacement vector, which indicates the movement of landmark $\mathbf{x}_i$ due to the brain shift process, can be calculated as  $\mathbf{d}(\mathbf{x}_i)=\mathbf{x}^{\mathrm{affine}_i}_i-\mathbf{x}_i$. where $\mathbf{d}=[d_x,d_y,d_z]$. 


	\vspace{-2mm}
\subsubsection{Dense deformation field} The goal of this step is to obtain a dense deformation field from a set of $N$ sparse landmark and their displacements $\mathcal{D}=\{ (\mathbf{x}_i,\mathbf{d}_i),i=1:N   \}$, where $\mathbf{d}_i=\mathbf{d}(\mathbf{x}_i)$ is modeled as a observation of displacements. 

In the GP model, let $\mathbf{d}(\mathbf{x})$ be the displacement vector for the voxel at location $\mathbf{x}$. thus it has a prior distribution as $d(\mathbf{x})\sim \mathrm{GP}(\mathrm{m}(\mathbf{x}),\mathrm{k}(\mathbf{x},\mathbf{x}'))$, where $\mathrm{m}(\mathbf{x})$ is the mean function, which usually is set to 0, and the GP kernel $\mathbf{k}(\mathbf{x},\mathbf{x}')$ represent the spatial correlation of displacement vectors.

By the model assumption, all displacement vectors follow a joint Gaussian distribution $p(\mathbf{d}\mid \mathbf{X})=\mathcal{N} (\mathbf{d}\mid \mathbf{\mu},\mathbf{K}) $, where $K_{ij}=\mathbf{k}(\mathbf{x},\mathbf{x}')$ and $\mathbf{\mu} = (\mathrm{m}(\mathbf{x}_1) ,...,\mathrm{m}(\mathbf{x}_N)) $. As a result, the displacement vectors $\mathbf{d}$ for known landmarks and $N_*$ unknown displacement vectors $\mathbf{d}_*$ at location $\mathbf{X}_*$, which we want to predict, have the following relationship:
	\vspace{-2mm}
\def\A{
	\begin{pmatrix}
		\mathbf{d}\\
	    \mathbf{d}_* \\
\end{pmatrix}}

\def\B{
	\begin{pmatrix}
		\mathbf{\mu} \\
		\mathbf{\mu}_*\\
	\end{pmatrix}}

\def\C{
	\begin{pmatrix}
		\mathbf{K} &	  \mathbf{K}_*\\
		\mathbf{K}^T_* &	  \mathbf{K}_{**}.\\
	\end{pmatrix}}
\begin{equation}
\A \sim \left(\B , \C\right)_.
\end{equation}

In Equation (1), $\mathbf{K}=\mathrm{k}(\mathbf{X},\mathbf{X})$ is the $N\times N$ matrix, $\mathbf{K}_*=\mathrm{k}(\mathbf{X},\mathbf{X_*})$ is the $N \times N_*$ matrix, and $\mathbf{K_{**}}=\mathrm{k}(\mathbf{X_*},\mathbf{X_*})$ is the $N_* \times N_*$ matrix. 
The mean $\mu_*=[\mu_{*x},\mu_{*y},\mu_{*z}]$ represents values of voxel-wise displacement vectors and can be estimated from the posterior Gaussian distribution $p(\mathbf{d}_*\mid \mathbf{X_*},\mathbf{X},\mathbf{d})= \mathcal{N}(\mathbf{d}_*\mid \mu_*,\Sigma_*)$ as 

\begin{equation}
\mu_*= \mu(\mathbf{X_*})+\mathbf{K}^T_*\mathbf{K}^{-1}(\mathbf{d}-\mu(\mathbf{X})).
\end{equation}

Given $\mu(\mathbf{X})= \mu(\mathbf{X_*})=0$, we can obtain the dense deformation field for the preUS image by assigning $\mu_{*x}$,$\mu_{*y}$,$\mu_{*z}$ to $\mathbf{d}_x$, $\mathbf{d}_y$ and $\mathbf{d}_z$, respectively.

\vspace{-4mm}
\subsubsection{Active registration} Automatic approaches may have difficulty in the preop-to-intraop image registration, especially for areas near the tumor resection site. Another advantage of the GP framework is the possibility of incorporate user expertise to further improve the registration result. 

From Equation (1), we can also compute the covariance matrix of the posterior Gaussian $p(\mathbf{d}_*\mid \mathbf{X_*},\mathbf{X},\mathbf{d})$ as
\begin{equation}
\Sigma_*= \mathbf{K}_{**}-\mathbf{K}^T_*\mathbf{K}^{-1}\mathbf{K}_*.
\end{equation}
Entries on the diagonal of $\Sigma_*$ are the marginal variances of predicted values. They can be used as an uncertainty measure to indicates the confidence in the estimated transformation parameters.

If users are not satisfied by the FBR alignment result, they could manually, guided by the image context and visualization of registration uncertainty, add new corresponding pairs of key-points to drive the GP towards better results.
 
\vspace{-4mm}
\subsection{GP kernel estimation}

The performance of GP registration depends exclusively on the suitability of the chosen kernels and its parameters. In this study, we explore two schemes for the kernel estimation: Variograms and discrete grid search.

\vspace{-4mm}
\subsubsection{Variograms} While being used extensively in geostatistics to characterize the spatial dependence of a stochastic process \cite{VarioBook}, variograms have not yet received much attention in the medical imaging field. Although GP regression for medical image registration, and variograms, were described in \cite{Warfield}, neither quantitative results, nor estimation of the posterior uncertainty were provided.

In the GP registration context, $\mathbf{d}(\mathbf{x})$ is modelled as a random quantity, variograms can measure the extent of pairwise spatial correlation between displacement vectors with respect to their distance, and in advance give insight into choosing the suitable GP kernel.

In practice, we estimate the empirical variogram of landmarks' displacement vectors as
\vspace{-3mm}
\begin{equation} 
\begin{aligned}
\hat{\gamma}(h\pm\delta)  
&:=\frac{1}{2|N(h\pm\delta)|}\sum_{(i,j)\in{N(h\pm\delta)}}^{}\norm{\mathbf{d}(\mathbf{x}_i)-\mathbf{d}(\mathbf{x}_j)}^2.
&
\end{aligned}
\end{equation}

For the norm term $\norm{\mathbf{d}(\mathbf{x}_i)-\mathbf{d}(\mathbf{x}_j)}$, we compute its 3 components $d_x$ $d_y$ $d_z$ and construct 3 variograms, respectively. As shown in Fig.3(a), for displacement vectors $\mathbf{d}(\mathbf{x_1})$ and $\mathbf{d}(\mathbf{x_2})$, $\norm{d_x(\mathbf{x}_2)-d_x(\mathbf{x}_1)}$ is the vector difference with respect to the $\mathbf{x}$ axis, etc. $h$ represents the distance between two displacement vectors.

\begin{figure}[t]
	\centering
	\includegraphics[height=3.8cm]{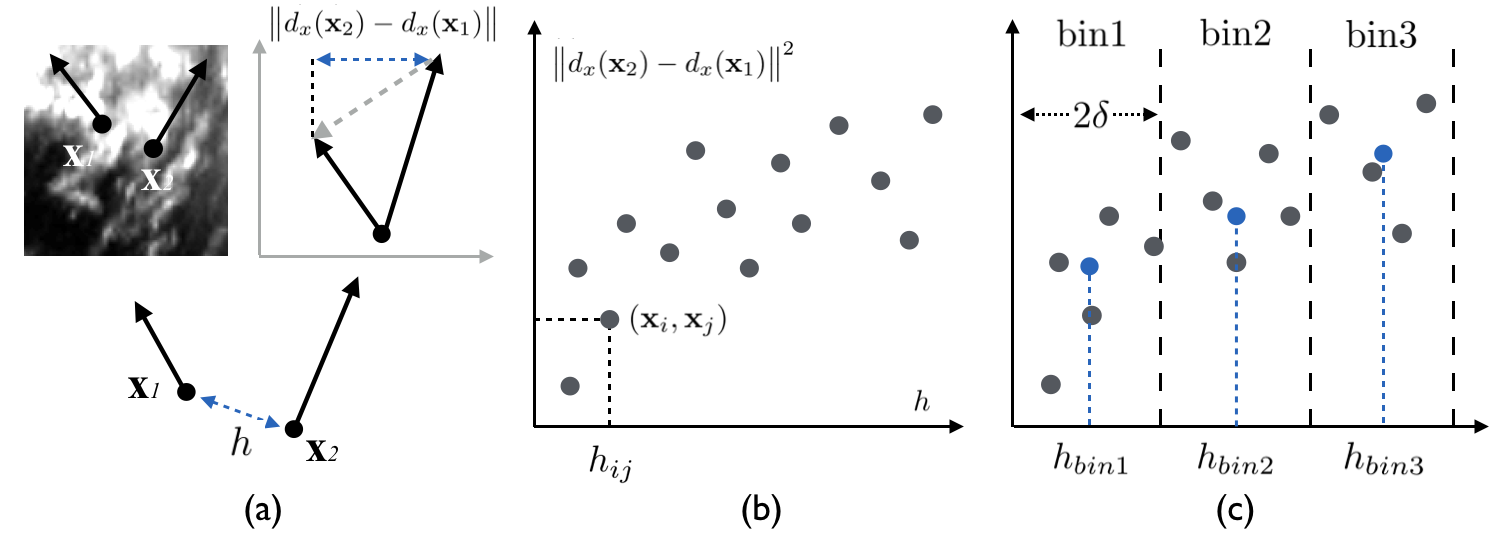}
	\vspace{-3mm}
	\caption{(a) $\norm{d_x(\mathbf{x}_2)-d_x(\mathbf{x}_1)}$  and $h$; (b) Empirical variogram cloud; (c) Variogram cloud divided into bins with their means marked as blue.}
	\label{fig:correct1}
	\vspace{-5mm}
\end{figure}

To construct an empirical variogram, the first step is to make a variogram cloud by plotting $\norm{d(\mathbf{x}_2)-d(\mathbf{x}_1)}^2$ and  $h_{ij}$ for all displacement pairs. Next, we introduce a variable $\delta$, and divide the variogram cloud into bins with a bin width setting to 2$\delta$. Lastly, the mean of each bin is calculated and further plotted with the mean distance of that bin to form an empirical variogram. Fig.4(a) shows an empirical variogram of a real US image pair that has 71 landmarks.

The empirical variogram only consists of value differences at a finite set of discrete distances, whereas the GP kernels are continuous for all $h$. Therefore, the next step is to fit a smooth curve to the empirical values and derive the kernel function from that fitted curve. 

	\vspace{-6mm}
\begin{figure}[H]
	\centering
	\includegraphics[height=3.1cm]{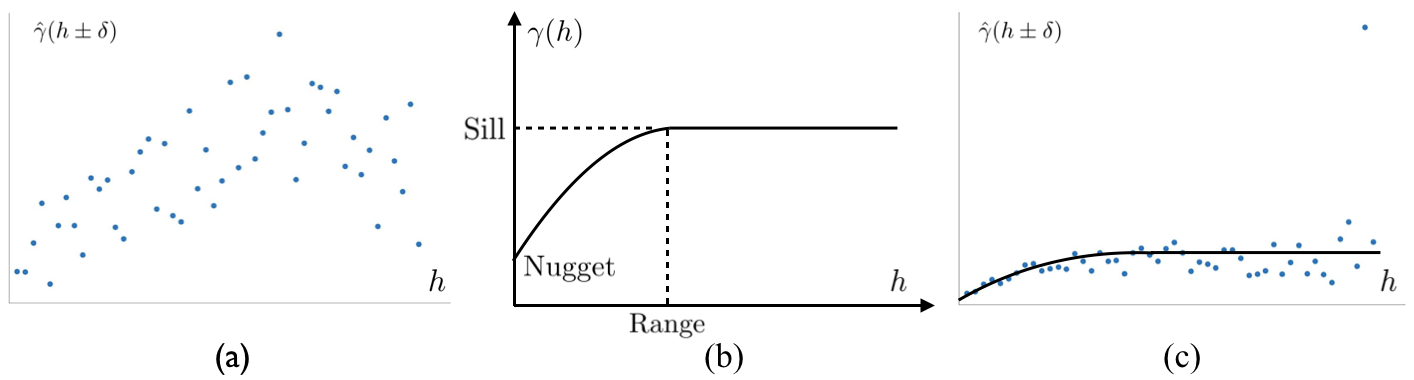}
	\vspace{-3mm}
	\caption{(a) X-axis empirical variogram of a US images pair;(b) Sill, range and nugget; (c) Fitting a continuous model to an empirical variogram.}
	\label{fig:correct1}
	\vspace{-6mm}
\end{figure}

Fig.4(b) is an example of a fitted curve. The curve is commonly describe by the following characteristics:
\vspace{-2mm}
\begin{labeling}{Parameters}
	\item [Nugget] The non-zero value at $h=0$.
	\item [Sill] The value at which the curve reaches its maximum.
	\item [Range] The value of distance $h$ where the sill is reached. 
\end{labeling}
\vspace{-1mm}
Conventionally, displacement vectors that are separated by distances further than the range are considered uncorrelated \cite{VarioBook}. 

In general, the curve must have a mathematical expression that can describe the variances of a random process. Practically, the choice is limited to a few options, such as exponential and Gaussian models. For instance, the Gaussian variogram function is
\vspace{-6mm}
\begin{equation}
\gamma(h)=c_0+c\{1-exp(- \frac{h^2}{a}  )\}.
\end{equation}

In equation (5), $c_0$ is the nugget, $c=\mathrm{Sill}-c_0$, and $a$ is the model parameter. This function asymptotically approaches its sill, and has an effective range as $r'=\sqrt{3a}$.    

Fitting a  model to an empirical variogram is implemented in most geostatistis software. A popular choice is choosing several models that appear to have the right shape and use the one with smallest weighted squared error \cite{VarioBook}.
\vspace{-3mm}
\subsubsection{Discrete grid search} The variogram scheme often requires many landmarks to work well \cite{VarioBook}. For US pairs that have fewer landmarks, we predefine some kernel functions, and use cross validation in a discrete search for the model parameters.

\vspace{-3mm}
\section{Experiments}
\vspace{-2mm}

The experimental dataset consists of 6 preUS and postUS image pairs that were acquired on a BK Ultrasound 3000 system (BK Medical, Analogic Corporations, Peabody, USA) that is directly connected to the Brainlab VectorVision Sky neuronavigaton system (Brainlab, Munich, Germany) during surgery.

We use the mean euclidean distance between the predicted and ground truth of landmarks' coordinates, measured in $mm$, for the registration evaluation. Compared methods include: affine, thin-plate kernel FBR, variograms FBR and gaussian kernel FBR. For US pairs that have less than 50 landmarks, we use leave-one-out cross validation, otherwise 5-fold cross validation. All of compared methods can be finished within 10 minutes.

\vspace{-5mm}
\begin{table}[H]
	\centering
	\caption{Registration evaluation results (in $\mathit{mm}$)}
	\label{my-label}
	\begin{tabular}{|c|c|c|c|c|c|c|}
		\hline
		& Landmarks & Before Reg. & Affine & Thin-plate & Variograms & GaussianK 
		\\ \hline
		Patient 1 & 123       &     5.56$\pm$1.05        &    2.99$\pm$1.21     &  1.79$\pm$0.70   &   2.11$\pm$0.74    & 1.75$\pm$0.68   \\ \hline
		Patient 2 & 71        &    3.35$\pm$1.22         &   2.08$\pm$1.13     &  2.06$\pm$1.18   &   2.06$\pm$1.12    &  1.97$\pm$1.05  \\ \hline
		Patient 3 & 49        &     2.48$\pm$1.56        &  1.93$\pm$1.75      & 1.25$\pm$1.95     &    n/a    &  1.23$\pm$1.77  \\ \hline
		Patient 4 & 12        &     4.40$\pm$1.79        & 3.06$\pm$2.35       &  1.45$\pm$1.99   &   n/a    &  1.42$\pm$2.04   \\ \hline
		Patient 5 & 64        &      2.91$\pm$1.33       &   1.86$\pm$1.24     &   1.29$\pm$1.17  & n/a   &  1.33$\pm$1.40  \\ \hline
		Patient 6 & 98        &    3.29$\pm$1.09        &  2.12$\pm$1.16      &  2.02$\pm$1.21   &  2.05$\pm$1.40    &  1.96$\pm$1.38  \\ \hline
	\end{tabular}
\end{table}

\vspace{-6mm}

In addition, we demonstrate the preliminary result of active registration. As shown in Fig.5, (a) is the registered source preUS image, (b) is the target postUS image. Noticing that the FBR does not well align the tumor boundary due to lacking of landmarks. In the active registration step, a user manually added 3 new key-point pairs based on the image context and a color mapping of registration uncertainty. By visual inspection, we can see the alignment of tumor boundary substantially improved. 

\begin{figure}[t]
	\centering
	\includegraphics[height=3.4cm]{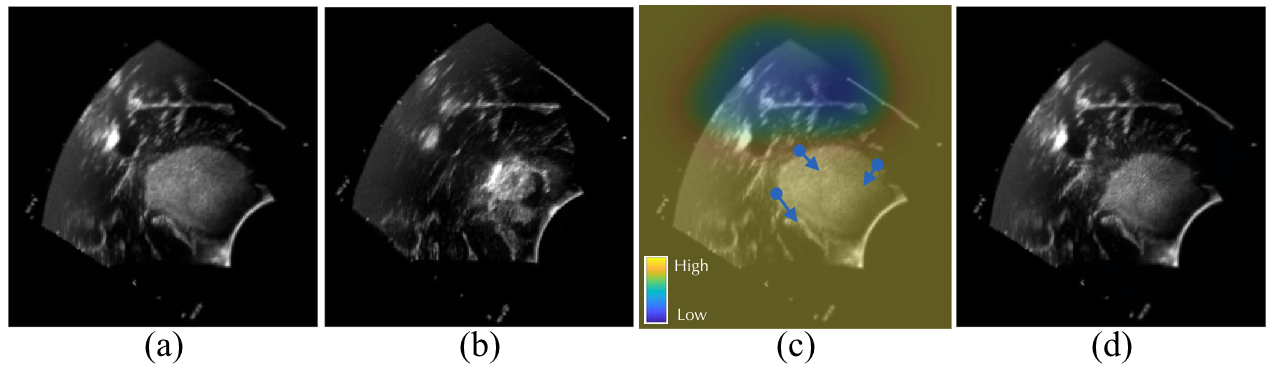}
	\vspace{-3mm}
	\caption{(a) FR result of the preUS image; (b) PostUS image; (c) Overlaying the visualization of uncertainty on the preUS image;(d) Improved registration result.}
	\label{fig:correct1}
	\vspace{-3mm}
\end{figure}

\vspace{-4mm}
\section{Conclusion}
\vspace{-3mm}

We proposed a novel feature-based active registration framework to compensate for the brain shift. We believe this framework has the potential to be eventually applied in the operating room. Future work includes exploring non-isotropic variograms and other advanced schemes for GP kernel estimation. Implementing our framework into clinical software, such as 3D slicer, is also of interest.

\end{document}